\documentclass{article}

\usepackage[preprint]{neurips_2024}

\usepackage[utf8]{inputenc} 
\usepackage[T1]{fontenc}    
\usepackage{hyperref}       
\usepackage{url}            
\usepackage{booktabs}       
\usepackage{amsfonts}       
\usepackage{nicefrac}       
\usepackage{microtype}      
\usepackage{xcolor}         
\usepackage{graphicx}
\usepackage{amsmath}
\usepackage[normalem]{ulem}
\usepackage{caption}
\usepackage{subcaption} 
\usepackage[misc]{ifsym}

\title{DreamPainter: Image Background Inpainting for E-commerce Scenarios}

\author{
  Sijie Zhao \quad Jing Cheng$^{~\textrm{\dag}}$ \quad Yaoyao Wu \quad Hao Xu$^{~\textrm{\Letter}}$ \quad Shaohui Jiao$^{~\textrm{\Letter}}$ \\ \\
  Bytedance 
}

\begin{document}

\maketitle

\let\thefootnote\relax\footnotetext{
\raggedright
${^{~\textrm{\dag}}}$~Project Lead \\
${^{~\textrm{\Letter}}}$~Corresponding Author
}

\begin{abstract}


Although diffusion-based image genenation has been widely explored and applied, background generation tasks in e-commerce scenarios still face significant challenges. The first challenge is to ensure that the generated products are consistent with the given product inputs while maintaining a reasonable spatial arrangement, harmonious shadows, and reflections between foreground products and backgrounds. Existing inpainting methods fail to address this due to the lack of domain-specific data. The second challenge involves the limitation of relying solely on text prompts for image control, as effective integrating visual information to achieve precise control in inpainting tasks remains underexplored. To address these challenges, we introduce DreamEcom-400K, a high-quality e-commerce dataset containing accurate product instance masks, background reference images, text prompts, and aesthetically pleasing product images. Based on this dataset, we propose DreamPainter, a novel framework that not only utilizes text prompts for control but also flexibly incorporates reference image information as an additional control signal. Extensive experiments demonstrate that our approach significantly outperforms state-of-the-art methods, maintaining high product consistency while effectively integrating both text prompt and reference image information.

\end{abstract}

\section{Introduction}

Diffusion-based image generation techniques ~\cite{rombach2021highresolution, podell2023sdxl,  esser2024scaling, zheng2024cogview3, flux2024, gao2025seedream} have experienced exponential growth in recent years, revolutionizing downstream tasks such as image editing~\cite{brooks2023instructpix2pix, tan2024ominicontrol, ge2024seed, wu2025less, deng2025bagel} and conditional-guided generation~\cite{mou2023t2i, zhang2023adding, peng2024controlnext}. Among these, Image Inpainting~\cite{suvorov2021resolution, lugmayr2022repaint, zhuang2024task, ju2024brushnet}, aiming to reconstruct coherent content in masked regions while preserving unmasked pixels, has emerged as a critical capability with broad applications in image restoration, object removal or insertion, background generation. Early inpainting approaches~\cite{suvorov2021resolution, lugmayr2022repaint} typically fine-tuned text-to-image base models on datasets with randomly generated masks, which often led to ambiguous object boundaries and unintended artifacts in generated regions due to the lack of structural guidance in mask definitions. Subsequent advancements~\cite{ju2024brushnet, zhuang2024task, fluxfill2024} introduced instance-aware masks to explicitly separate foreground objects from backgrounds, improving boundary consistency and reducing intrusion of irrelevant content .

In e-commerce scenarios, Image Inpainting holds particular promise for enhancing product presentation by retaining the foreground merchandise and generating aesthetically pleasing backgrounds. However, existing inpainting frameworks face two fundamental challenges in this specialized domain. First, most methods~\cite{fluxfill2024, fluxcontrolnet2024} are trained on generic datasets that lack of high-fidelity e-commerce-specific imagery, resulting in backgrounds that suffer from poor resolution, inconsistent lighting, or mismatched stylistic elements when applied to product images. Second, relying solely on textual descriptions to guide background generation proves insufficient when prompts involve abstract concepts. Although text offers semantic guidance, it often fails to capture nuanced visual aesthetics (e.g., subtle color gradients, material textures, or compositional styles) that are critical for professional design intent. This limitation becomes especially pronounced when creators seek to translate abstract artistic visions into precise visual outputs, highlighting the need for more effective conditioning mechanisms that go beyond textual inputs.

Integrating visual guidance has emerged as a pivotal research direction, demonstrating growing significance for bridging semantic understanding and visual representation. ControlNet~\cite{zhang2023adding} pioneered a modular architecture that incorporates auxiliary inputs like depth, canny, and pose through dedicated conditioning modules, enabling fine-grained control over synthetic outputs. Concurrently, IP-Adapter~\cite{ye2023ip} adopted a lightweight strategy by leveraging CLIP-style encoders to extract semantic-rich visual features, which are then fed as conditional signals to steer the generation process. The recent works~\cite{tan2024ominicontrol, wu2025less} advanced this paradigm by directly encoding visual cues into the latent space, utilizing attention mechanisms to dynamically modulate feature interactions during synthesis. Although these innovations have significantly improved the controllability of image generation tasks, their application to image inpainting remains largely unexplored.

The unique demands of e-commerce image generation, where product clarity, brand consistency, and visual appeal directly impact consumer engagement, require a tailored approach that addresses both data quality and conditional guidance. Unlike generic inpainting tasks, e-commerce applications require tight alignment between generated backgrounds and the intrinsic attributes of foreground products (e.g., material, color, usage context), as well as the ability to incorporate implicit design preferences that may be difficult to articulate through text alone. Failing to meet these requirements can lead to disjointed visual experiences, which undermines the marketing effectiveness of product displays.

\begin{figure}[t]
	\centering
	\includegraphics[width=1.0\linewidth]{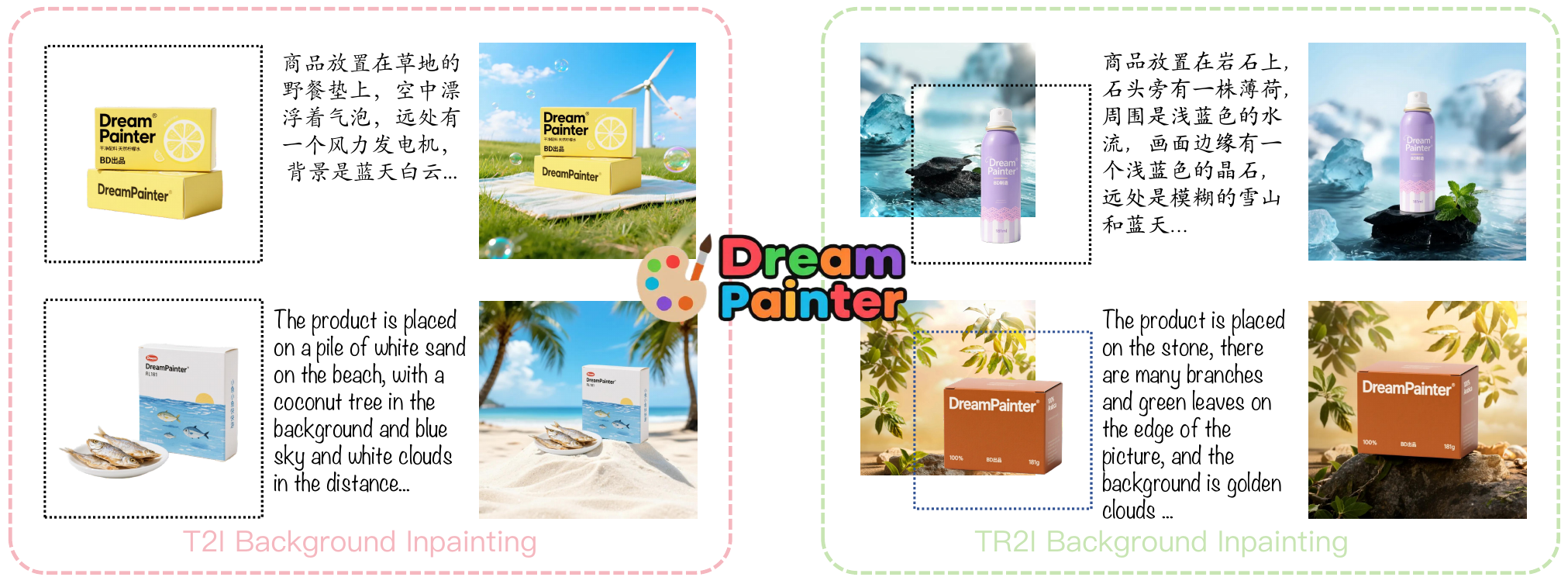}
    \caption{Overview of DreamPainter, which can not only perform background inpainting based on text prompts (T2I), but also combine reference image information (TR2I).}
\label{fig:overview}
\vspace{-5pt}
\end{figure}

To address these challenges, we introduce DreamPainter, a framework for e-commerce product background generation. As shown in Fig.~\ref{fig:overview}, our approach leverages both textual descriptions and visual reference signals to provide multi-modal conditioning, enabling precise control over background semantics and style while preserving product fidelity. To achieve this goal, we build a high-quality large-scale dataset, DreamEcom-400K, specifically curated for e-commerce image manipulation, which captures diverse product categories, backgrounds, and professional photography standards. By bridging the gap between generic inpainting techniques and domain-specific requirements, our work aims to establish a new benchmark for generating contextually appropriate, artistically aligned backgrounds that empower e-commerce platforms and designers to create compelling visual content.

\section{Related Work}

\paragraph{Image Inpainting}

Image inpainting, a critical technique in image generation, focuses on restoring missing image regions by leveraging existing contextual information, with broad applications in object removal, insertion, replacement, and background generation. Early approaches~\cite{zhao2021large, suvorov2021resolution, wan2021high} based on generative adversarial networks~\cite{goodfellow2020generative} used randomly shaped masks to create training data, but struggled to synthesize visually coherent content with novel semantic information. The advent of diffusion models~\cite{rombach2021highresolution, podell2023sdxl, esser2024scaling} revolutionized image generation, significantly enhancing inpainting performance for tasks requiring new semantic content. However, the continued use of random masking in these methods often blurred object boundaries, leading to the generation of unwanted artifacts at region interfaces. Recent advancements in object detection~\cite{liu2023grounding, zhao2024omdet} and segmentation~\cite{kirillov2023segment, ravi2024sam} enabled the extraction of precise instance masks. This innovation reduced artifact occurrences by preserving clear object boundaries during the inpainting process. In the e-commerce domain, where the goal is to generate aesthetically pleasing backgrounds for product foregrounds, traditional methods fall short due to their lack of domain-specific knowledge. Although some recent works~\cite{sdxlcontrolnet2024, fluxcontrolnet2024} have focused on e-commerce scenarios, their reliance on low-aesthetic training data limits the quality of generated results. To address these gaps, our research utilizes the state-of-the-art Seedream3~\cite{gao2025seedream} to construct the DreamEcom-400k dataset, incorporating 400k high-quality e-commerce images for training. We also introduce DreamPainter, a robust inpainting framework built on open source CogView4~\cite{zheng2024cogview3}, specifically tailored to meet the demanding requirements of e-commerce image generation.

\paragraph{Visual Guidance Generation}
In e-commerce image synthesis, where precise control over subtle visual attributes (e.g., nuanced color variations, element arrangements) is critical, relying solely on textual prompts often proves insufficient. These scenarios demand more granular alignment between desired visual properties and generated outputs.
Research on the integration of diverse visual cues into image generation has made substantial progress, with methods like ControlNet~\cite{zhang2023adding} and T2I-Adapter~\cite{mou2024t2i} leading the way by conditioning generation on explicit visual modalities such as pose, canny and depth. Subsequently, IP-Adapter~\cite{ye2023ip} introduced a CLIP-style encoder to extract semantic-rich features from reference images, enabling more context-aware generation by leveraging high-level visual semantics rather than low-level structural cues. Recent advances~\cite{tan2024ominicontrol, wu2025less} have shifted the focus to encoding source images directly into the latent space of generative models, using attention mechanisms to selectively propagate spatial or semantic control signals, an approach that achieves finer-grained manipulation over output details.
In addition, unified multimodal frameworks~\cite{ge2024seed, wang2024emu3, deng2025bagel} have emerged that facilitate the natural integration of visual guidance through conversational interfaces or contextual prompting, which bridges the gap between diverse input modalities and generation tasks. By treating visual guidance as an essential control dimension alongside text, our work addresses the stringent requirements of e-commerce applications, where pixel-level accuracy and semantic consistency are paramount.

\section{Dataset}

\begin{figure}[t]
	\centering
	\includegraphics[width=1.0\linewidth]{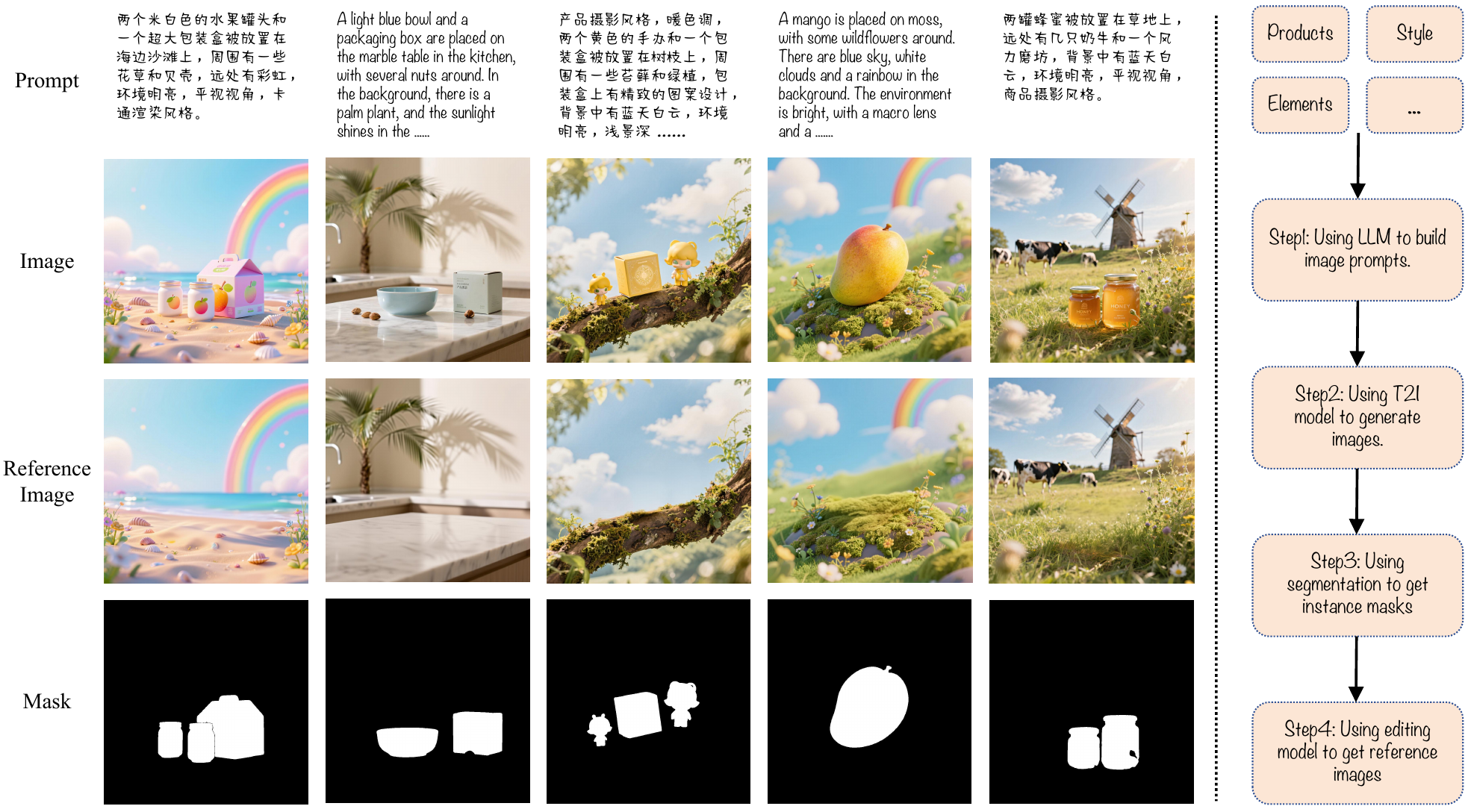}
    \caption{Data samples from DreamEcom-400k (left) and construction pipeline (right).}
\label{fig:datapipe}
\vspace{-5pt}
\end{figure}


To address the challenges in e-commerce image generation, we design a systematic pipeline to produce high-quality image datasets as shown in Fig.~\ref{fig:datapipe} (right), Which includes the following steps:

\textbf{Step 1: Using LLM to build image prompts.} The visual content is decomposed into five semantic components: (1) products, encompassing attributes such as categories, quantities, and colors; (2) display stands, defined as objects in direct contact with commodities; (3) environmental elements, referring to surrounding contextual features; (4) backgrounds, representing distant scenery; and (5) style, such as photographic styles or 3D rendering aesthetics. By setting attribute ranges for each component, controlling sampling ratios, and performing random combinatorial operations, we leverage LLM to expand these structured inputs into semantically rich image prompts.

\textbf{Step2: Using T2I model to generate images.} We utilize the state-of-the-art text-to-image (T2I) model Seedream3~\cite{gao2025seedream} to generate commodity images based on the constructed prompts. Empirical observations demonstrate that Seedream3 produces images of exceptionally high quality, eliminating the need for additional post-processing filters during the generation stage.

\textbf{Step3: Using segmentation method to get instance masks.} For instance segmentation, we adopt a hybrid approach combining GroundingDINO~\cite{liu2023grounding} and the SAM~\cite{ravi2024sam}. By configuring multiple detection categories for each image during inference, this pipeline ensures comprehensive and accurate segmentation of commodity instances.

\textbf{Step4: Using editing model to get reference images.} To create background reference images, we employ the unified multimodal model BAGLE~\cite{deng2025bagel} for semantic editing. Unlike traditional inpainting methods~\cite{suvorov2021resolution, ju2024brushnet, zhuang2024task} that often introduce extraneous objects in masked regions or fail to remove shadows/reflections effectively, BAGLE enables direct execution of editing instructions (e.g., "remove the product and its shadows/reflections") with high fidelity. This capability addresses the limitations of conventional approaches, providing clean background reference images by seamlessly erasing target commodities and their associated visual effects.

Through this integrated workflow, we have developed the DreamEcom-400k dataset, comprising 400k high-quality and diverse product samples. As shown in Fig.~\ref{fig:datapipe} (left), each sample includes a commodity image, bilingual (Chinese-English) prompts, instance masks, and background reference images, establishing a robust foundation for background generation in e-commerce scenarios.

\section{Methodology}

In this section, we present the details of our proposed DreamPainter. We first elaborate on the system framework, followed by the two-stage training strategy, and finally discuss the flexible control mechanism for reference images.

\subsection{System Framework}

The core of DreamPainter lies in transforming a text-to-image (T2I) model into a background inpainting model with flexible reference guidance. We define the foreground image as \( I_f \), the mask as \( M \), the target image as \( I_{\text{tgt}} \), and the background reference image as \( I_{\text{ref}} \). Using a VAE encoder, we encode \( I_{\text{tgt}} \) into \( Z_{\text{tgt}} \), \( I_f \) into \( Z_f \), and \( I_{\text{ref}} \) into \( Z_{\text{ref}} \). Let \( Z^t_{\text{tgt}} \) denote the noisy latent obtained by adding \( t \)-step noise to \( Z_{\text{tgt}} \). The text prompt is encoded into \( C \) via a text encoder, and \( M \) is downsampled to \( Z_m \) to align its spatial dimensions with \( Z_f \).
To adapt the T2I model for background inpainting, we consider two main strategies for integrating conditional information.

\textbf{Input-layer channel concatenation.} This strategy concatenates \( Z_m \), \( Z_f \), and \( Z^t_{\text{tgt}} \) along the channel dimension, followed by concatenating the result with \( C \) along the token (spatial) dimension as the input to the first layer:
\begin{equation}
Z_1 = \mathcal{C}_s\left( C,\ \mathcal{C}_c\left( Z^t_{\text{tgt}},\ Z_f,\ Z_m \right) \right)
\end{equation}
where \( \mathcal{C}_c(\cdot) \) denotes channel-wise concatenation, and \( \mathcal{C}_s(\cdot) \) denotes spatial dimension concatenation. This approach only modifies the channels of input layer, allowing for either full fine-tuning or lightweight LoRA-based adaptation during training.

\textbf{Auxiliary control branch.} This strategy introduces a separate control branch to process \( Z_f \), \( Z_m \) , \( Z^t_{\text{tgt}} \) and \( Z^t_{\text{tgt}} \), which is then integrated into the \( i \)-th layer of the original model:
\begin{equation}
Z_i = Z_i + \mathcal{G}_j\left( \Theta_j\left( Z^t_{\text{tgt}},\ Z_f,\ Z_m \right) \right)
\end{equation}
Here, \( \Theta_j(\cdot) \) represents the output of the \( j \)-th layer in the control branch, and \( \mathcal{G}_j \) is a linear gate layer initialized with zero weights. Although this method introduces additional parameters and computation due to the auxiliary branch, recent work on video generation \cite{jiang2025vace} have shown that such branches yield smaller training losses compared to input-layer adjustments. Thus, we adopt this control branch design and initialize its modules using weights from the original DiT architecture to stabilize training.

To inject reference image information into the generation process, we follow prior works \cite{tan2024ominicontrol,wu2025less} and concatenate \( C \), \( Z^t_{\text{tgt}} \), and \( Z_{\text{ref}} \) along the spatial dimension as input to the DiT blocks:
\begin{equation}
Z_1 = \mathcal{C}_s\left( c,\ Z^t_{\text{tgt}},\ Z_{\text{ref}} \right)
\end{equation}
We assign distinct positional encodings to preserve the 2D spatial relationship between \( Z^t_{\text{tgt}} \) and \( Z_{\text{ref}} \) avoid confusion. Specifically, given \( Z^t_{\text{tgt}} \) with height \( h \) and width \( w \), we retain the original positional indices for \( Z^t_{\text{tgt}} \) and assign new indices to \( Z_{\text{ref}} \) tokens at position \( (i,j) \):
\begin{equation}
(\hat{i},\ \hat{j}) = \left( h + i,\ w + j \right)
\end{equation}
This design mimics placing a reference image beside the target canvas, enabling the model to better leverage reference information. To adapt to the new input pattern, we incorporate learnable LoRA into the DiT blocks.

\begin{figure}[t]
	\centering
	\includegraphics[width=1.0\linewidth]{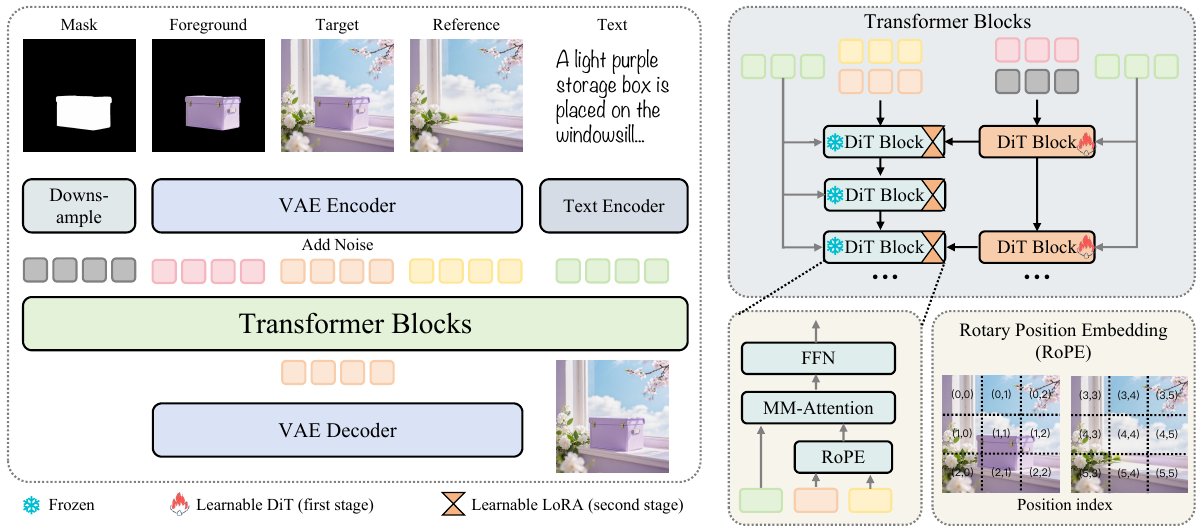}
    \caption{Framework of DreamPainter. We inject mask and foreground information through the control branch and update them in the first training stage. In the second training stage, we add learnable LoRA to the base model to process reference image information. We perform special positional encoding for reference images to avoid confusion.}
\label{fig:framework}
\vspace{-5pt}
\end{figure}

\subsection{Two-Stage Training Strategy}

Incorporating reference images as additional conditions simplifies the learning task compared to pure text-based control. To retain background-inpainting capability under text-only control, we propose a progressive two-stage training strategy.

\textbf{Stage 1: Text-only control.} We first train the model using only text prompts as control signals. During this stage, only the parameters of the control branch are optimized to focus on learning text-driven background inpainting.

\textbf{Stage 2: Reference-aware adaptation.} In the second stage, we freeze the control branch and original model parameters, then introduce learnable LoRA into the DiT blocks. To balance text and reference control, we randomly drop reference information with a high probability during training. This ensures the final model can handle both text-only and text-reference joint control scenarios. Additionally, we augment reference images with random spatial shifts to enhance the model’s ability to fuse foreground-background spatial relationships.

\subsection{Flexible Control of Reference Images}

In practical inference, users often desire generated backgrounds that resemble but are not identical to the reference. To balance similarity and randomness, we design two mechanisms to adjust reference influence.

 \textbf{LoRA scaling.} Since the LoRA parameters are optimized specifically for reference integration in stage 2, we can scale their output to modulate reference influence:
\begin{equation}
f_{\text{out}} = \phi\left( f_{\text{in}} \right) + s \cdot \text{LoRA}\left( f_{\text{in}} \right)
\end{equation}
where \( \phi(\cdot) \) denotes the linear layer in original DiT blocks, \( s \) is a scale coefficient, and \( \text{LoRA}(\cdot) \) is the corresponding LoRA layer.

\textbf{Attention modulation.} In the DiT attention mechanism, the attention score matrix is computed as \( \text{softmax}\left( \frac{\mathbf{QK}^\top}{\sqrt{d}} \right) \), representing the correlation between different tokens. We modulate the scores between \( Z^t_{\text{tgt}} \), \( C \) and \( Z_{\text{ref}} \) using a scaling matrix \( \mathbf{S} \),
\begin{equation}
\text{Attention}\left( \left[ C,\ Z^t_{\text{tgt}},\ Z_{\text{ref}} \right] \right) = \mathbf{S} \odot \text{softmax}\left( \frac{\mathbf{QK}^\top}{\sqrt{d}} \right) \mathbf{V}
\end{equation}
where \( \odot \) denotes element-wise multiplication. Given \( C \) with length \( m \), \( Z^t_{\text{tgt}} \) with length \(n \), and \( Z_{\text{ref}} \) with length \( l \), \( \mathbf{S} \) is structured as:
\begin{equation}
S = \begin{bmatrix} 
\mathbf{1}_{m \times m} & \mathbf{1}_{m \times n} & s \cdot \mathbf{1}_{m \times l} \\ 
\mathbf{1}_{n \times m} & \mathbf{1}_{n \times n} & s \cdot \mathbf{1}_{n \times l} \\ 
s \cdot \mathbf{1}_{l \times m} & s \cdot \mathbf{1}_{l \times n} & \mathbf{1}_{l \times l} 
\end{bmatrix}
\end{equation}
By adjusting \( s \), users can flexibly control the trade-off between reference similarity and generative randomness.

In summary, DreamPainter integrates foreground and mask information via an auxiliary control branch, injects reference guidance through spatial concatenation with special positional encoding, and leverages a two-stage training strategy to balance text and reference control. The flexible scaling mechanisms further enable user-adjustable reference influence during inference.

\section{Experiments}

\subsection{Experimental Setups}

\textbf{Trainging Details.} DreamPainter is built upon CogView4~\cite{cogview4,zheng2024cogview3}, for its appropriate model size and dual-language (Chinese and English) support. The model was trained on our proposed DreamEcom-400k dataset. The training process was composed of two stages: (1) \textbf{Stage 1}: We introduced a control branch comprising 6 layers of DIT Blocks, fine-tuning only this branch while keeping other parameters frozen. A constant learning rate of 4e-5 was employed, and the training process consumed approximately 1k GPU hours. Training images were processed at resolutions of $800\times800$ and $1024\times1024$. (2) \textbf{Stage 2}: LoRA modules with a rank of 256 were added to all linear layers of the original branch. During training, reference images were incorporated at a sampling rate of 50\%. These references were randomly resized and cropped to the resolution of $512\times512$, and $800\times800$. A cosine annealing learning rate schedule was adopted, with a peak learning rate of 8e-5. This stage required approximately 700 GPU hours.

\textbf{Evaluation Details} We designed a specialized evaluation dataset based on real-world e-commerce scenarios and product images. The dataset consists of two subsets: (1) 164 samples for Text-to-Image (T2I) background inpainting, each accompanied by a prompt and a product image on a white background and a mask; (2) 122 samples for Text-and-Reference-to-Image (TR2I) background inpainting, with each entry including a product image, prompt, reference image, and mask.
The evaluation employed the following metrics: (1) \textbf{IR Score}~\cite{xu2023imagereward} and (2) \textbf{PickScore}~\cite{kirstain2023pick}, which jointly assess prompt adherence, visual appeal, and compositional harmony across the entire image. (3) \textbf{Object Consistency}, we utilize GroundDINO~\cite{liu2023grounding} and SAM~\cite{kirillov2023segment} for product detection and segmentation, we develop a formula to measure shape consistency: $1 - \frac{\sum(M_{\text{gen}} \notin M_{\text{gt}})}{\sum(M_{\text{gt}})}$. A value closer to 1 indicates that the product can be accurately blended on the generated background without artifacts. (4) \textbf{CLIP Score}~\cite{radford2021learning}: Evaluates semantic consistency between the generated image and the prompt, as well as between the generated image and the reference image. For reference image comparisons, we directly composite the product onto the reference image to prevent foreground omissions skewing similarity scores.

\subsection{Evaluation on T2I Inpainting}
We conducted comprehensive comparisons with several state-of-the-art open-source methods on the T2I Background Inpainting task. These methods include: (1) EcomXL-Inpainting~\cite{sdxlcontrolnet2024}, an inpainting model for e-commerce scenarios built upon SDXL~\cite{podell2023sdxl} and ControlNet~\cite{zhang2023adding}; (2) FLUX-Inpainting~\cite{fluxcontrolnet2024}, a general inpainting model based on FLUX~\cite{flux2024} and ControlNet~\cite{zhang2023adding}; (3) FLUX-Fill~\cite{fluxfill2024}, another FLUX-based general inpainting model; (4) OmniControl~\cite{tan2024ominicontrol}, a versatile editing model supporting inpainting mode; (4) UNO~\cite{wu2025less}, a model capable of editing up to four input images. During inference, products were centered within a $0.6 ~\text{height} \times 0.6~\text{width}$ region by resizing and padding, and evaluations were performed at a resolution of $1024\times 1024$.

\begin{figure}[htbp]
    \centering
    \begin{subtable}[b]{0.55\textwidth} 
        \centering
        \small
        \setlength{\tabcolsep}{3.5pt}
        \begin{tabular}{lcccc} 
            \toprule
            Method            & IR-Score       & PickScore      & \begin{tabular}[c]{@{}c@{}}Object \\consistency\end{tabular} & CLIP-T          \\ 
            \hline
            EcomXL-Inp.~\cite{sdxlcontrolnet2024}  & 0.459          & 0.097          & \uline{0.958}                                                & 0.335           \\
            FLUX-Inp.~\cite{fluxcontrolnet2024}   & 0.593          & 0.095          & 0.895                                                        & 0.331           \\
            FLUX-Fill~\cite{fluxfill2024}         & 0.705          & 0.097          & 0.938                                                        & 0.333           \\
            OmniControl~\cite{tan2024ominicontrol}       & \uline{0.714}  & \uline{0.120}  & 0.945                                                        & 0.329           \\
            UNO~\cite{wu2025less}               & 0.341          & \uline{0.120}  & 0.502                                                        & \textbf{0.345}  \\
            \textbf{Ours}     & \textbf{0.841} & \textbf{0.195} & \textbf{0.986}                                               & \uline{0.337}   \\
            \bottomrule
        \end{tabular}
        \caption{Quantitative evaluation on T2I Inpainting.}
        \label{tab:t2i_results}
    \end{subtable}
    \hfill
    \begin{subfigure}[b]{0.35\textwidth} 
        \centering
        \includegraphics[width=\textwidth]{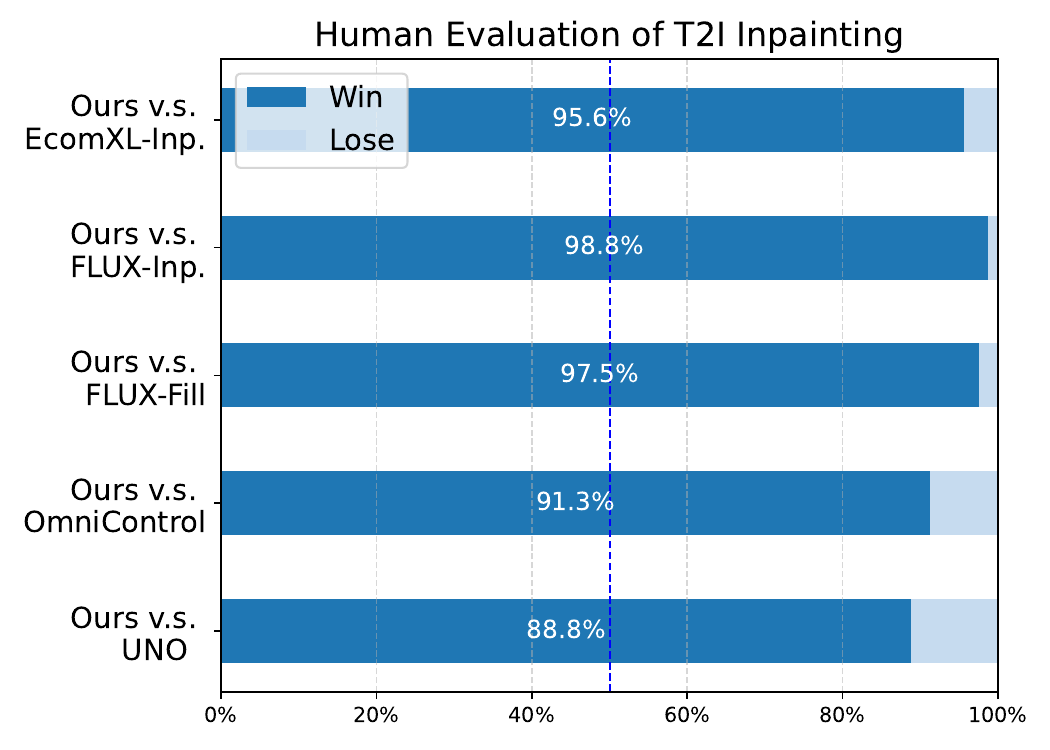}
        \caption{Human evalution on T2I inpainting.}
        \label{fig:t2i_humaneval}
    \end{subfigure}
    \caption*{ }
\end{figure}

\textbf{Quantitative Evaluation} Results from various quantitative metrics are presented in Tab.~\ref{tab:t2i_results}, where the best performance is bolded and the second-best underlined. Our method achieved leading scores in both the IR-Score (0.841 vs. 0.714) and the PickScore (0.195 vs. 0.12), significantly outperforming the second-ranked methods. In the results of Object Consistency, our model attained a remarkable 0.986, indicating minimal alterations to foreground objects. In contrast, UNO scored only 0.502, suggesting frequent distortions in product shape and quantity.

\textbf{Human Evaluation.} To better align with human preferences, we conducted pairwise comparisons in which human evaluators were asked to select the most suitable and highest quality result between our method and each baseline. Four evaluators participated to ensure reliability. As shown in Fig.~\ref{fig:t2i_humaneval}, our method achieved win rates of 95.6\%, 98.8\%, 97.5\%, 91.3\%, and 88.8\% against EcomXL-Inpainting, FLUX-Inpainting, FLUX-Fill, OmniControl, and UNO, respectively.

\begin{figure}[t]
	\centering
	\includegraphics[width=1.0\linewidth]{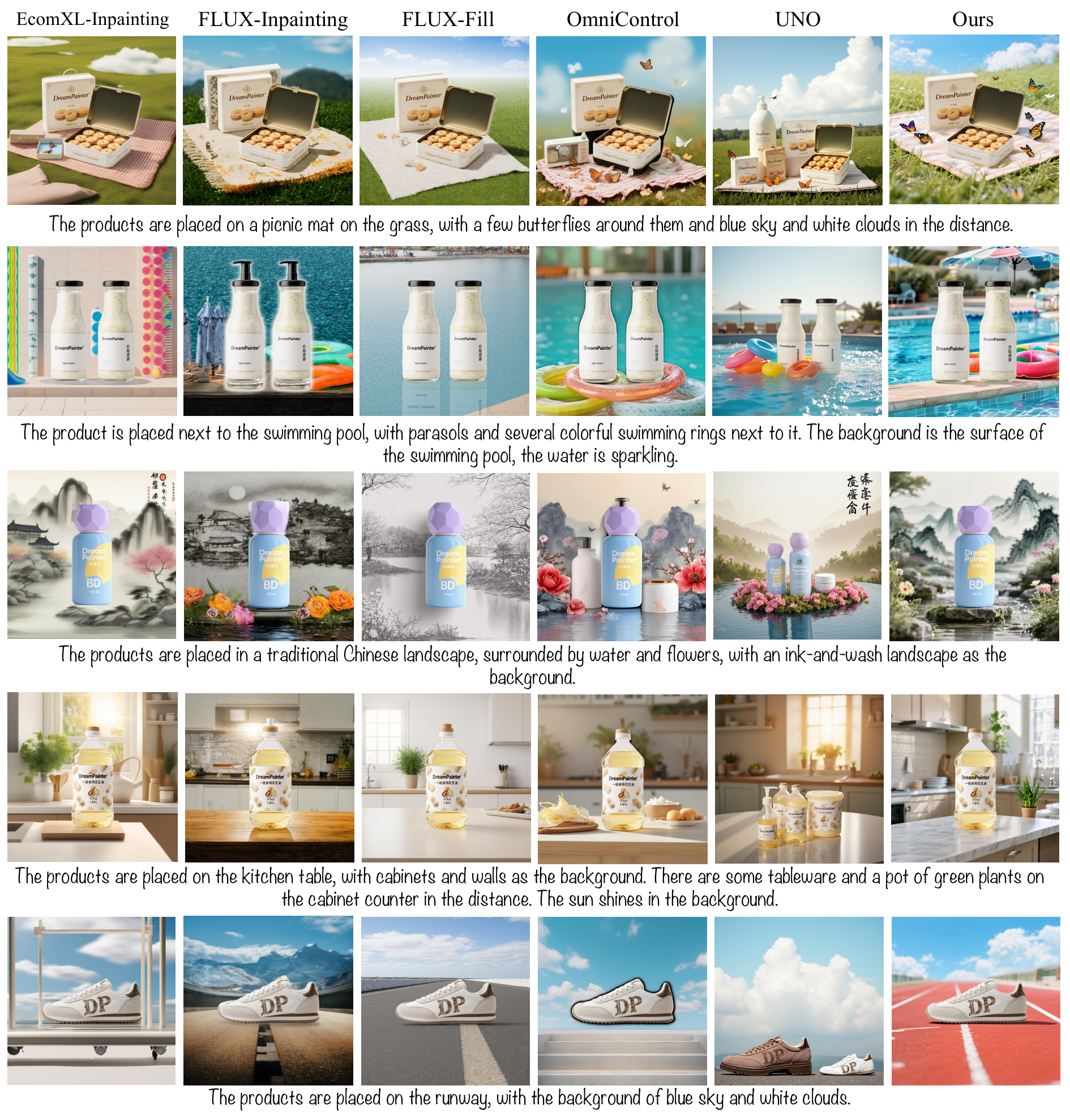}
    \caption{Qualitative comparison of T2I Inpainting between different methods.}
\label{fig:t2i_comparison}
\vspace{-5pt}
\end{figure}

\textbf{Qualitative Evaluation.} Visual comparisons in Fig.~\ref{fig:t2i_comparison} illustrate the superior performance of our method across diverse scenarios. While our approach consistently generated high-quality backgrounds while preserving product integrity, baselines exhibited notable flaws. FLUX-Inpainting and FLUX-Fill often introduced noisy backgrounds due to limited domain data, OmniControl produced dark artifacts around product edges, and UNO struggled to maintain object consistency. These results underscore the effectiveness of our method in achieving both fidelity to foreground objects and realism in background generation.

\subsection{Evaluation on TR2I Inpainting}

\begin{figure}[htbp]
    \centering
    \begin{subtable}[b]{0.45\textwidth} 
        \centering
        \setlength{\tabcolsep}{3.5pt}
        \begin{tabular}{lcccc} 
        \toprule
        Method        & IR-Score       & PickScore      & \begin{tabular}[c]{@{}c@{}}Object \\consistency\end{tabular} & CLIP-I          \\ 
        \hline
        UNO~\cite{wu2025less}           & 0.178          & 0.206          & 0.593                                                        & 0.880           \\
        BAGEL~\cite{deng2025bagel}         & \uline{0.393}  & \uline{0.233}  & \uline{0.793}                                                & \uline{0.897}   \\
        \textbf{Ours} & \textbf{0.932} & \textbf{0.506} & \textbf{0.992}                                               & \textbf{0.973}  \\
        \bottomrule
        \end{tabular}
        \caption{Quantitative evaluation on TR2I Inpainting.}
        \label{tab:tr2i_results}
    \end{subtable}
    \hfill
    \begin{subfigure}[b]{0.4\textwidth} 
        \centering
        \includegraphics[width=\textwidth]{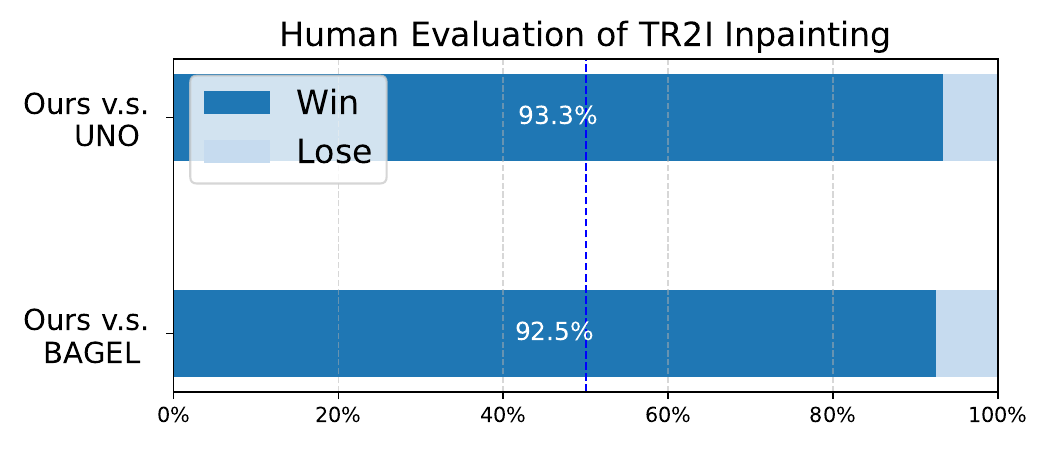}
        \caption{Human evaluation on TR2I inpainting.}
        \label{tab:tr2i_human_eval}
    \end{subfigure}
    \caption*{ } 
\end{figure}

\begin{figure}[t]
    
	\centering
	\includegraphics[width=1.0\linewidth]{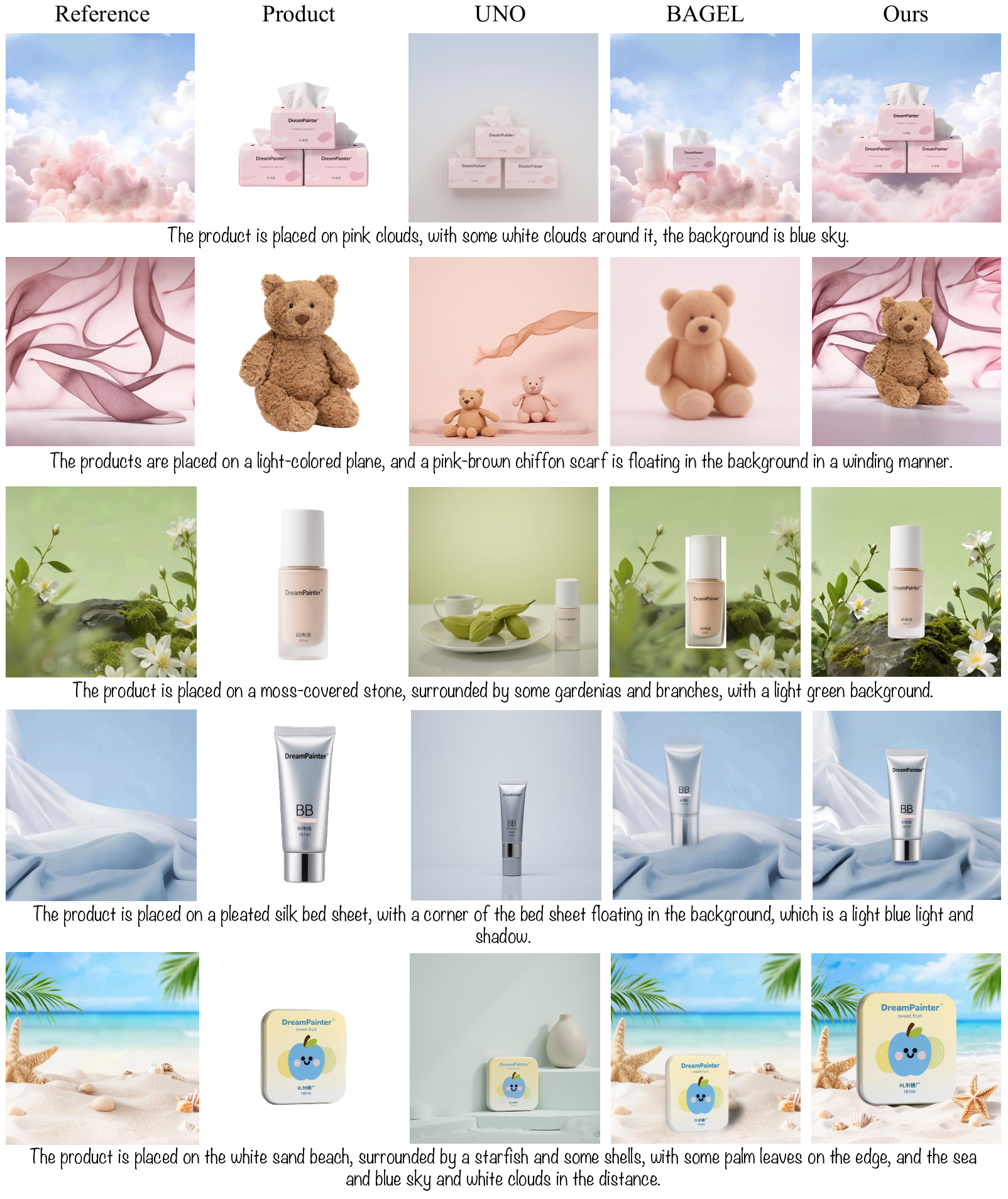}
    \caption{Qualitative comparison of TR2I Inpainting between different methods.}
\label{fig:tr2i_comparison}
\vspace{-5pt}
\end{figure}

Due to the lack of algorithms capable of incorporating reference image information for inpainting, we compare our method with state-of-the-art multi-image editing approaches: (1) UNO~\cite{wu2025less}, an FLUX-based model supporting up to four input images for editing, and (2) BAGEL~\cite{deng2025bagel}, a unified multimodal model enabling interleaved input of arbitrary images and text for comprehensive understanding and generation. Except for adding reference images, we maintain the same experimental settings as those in the T2I Inpainting task.

\textbf{Quantitative Evaluation.} Tab.~\ref{tab:tr2i_results} summarizes the quantitative results across multiple metrics. Our method significantly outperforms baselines in all indicators, with the best results bolded and second-best underlined. Notably, we achieve a 137.2\% lead over BAGEL (the second-best model) in IR-Score (0.932 vs. 0.393), a 125.8\% advantage in PickScore (0.506 vs. 0.233), a 25.1\% improvement in Object Consistency (0.992 vs. 0.793), and an 8.5\% lead in CLIP-I score (0.973 vs. 0.897).

\textbf{Human Evaluation.} To measure alignment with human preferences in the TR2I Inpainting task, we conduct a pairwise user study with the same setting in T2I Inpaitning task. As shown in Fig.~\ref{tab:tr2i_human_eval}, our method achieves win rates of 93.3\% and 92.5\% against UNO and BAGEL, respectively, indicating qualitative superiority in generating visually appealing and contextually appropriate outputs based on the reference images.

\textbf{Qualitative Evaluation.} Visual comparisons in Fig.~\ref{fig:tr2i_comparison} highlight the qualitative advantages of our approach. We generate high-fidelity product images where the object maintains strict consistency with the input and seamlessly integrates into the background. In contrast, baselines fail to produce stable, high-quality results, often resulting in inconsistent object representations or poor background coherence. We also demonstrate the visual effects of the background inpainting on out-of-domain tasks in the Appendix.~\ref{sec:out-of-domain}.

\subsection{Analyses}

\begin{figure}[t]
	\centering
	\includegraphics[width=1.0\linewidth]{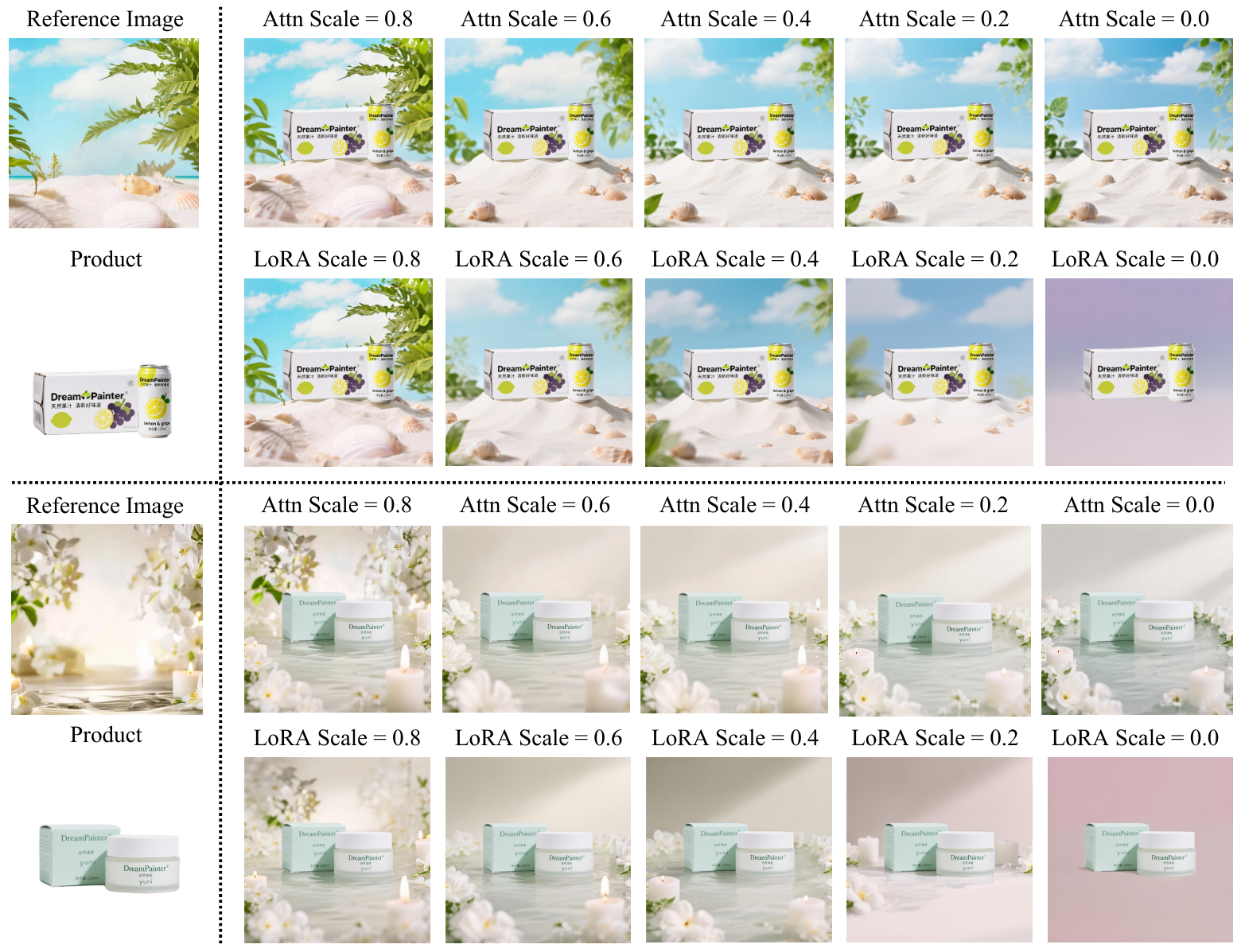}
    \caption{Qualitative analyses between Attention modulation and LoRA scaling.}
\label{fig:ablation}
\vspace{-5pt}
\end{figure}

\textbf{LoRA scaling vs Attention modulation.} We designed two methods to adjust the similarity between the generated images and the reference images. To assess those methods, we conducted the following experiments: (1) Fixing the LoRA scale at 1.0, we set the Attention scale to 0.8, 0.6, 0.4, 0.2, and 0.0 respectively; (2) Fixing the Attention scale at 1.0, we set the LoRA scale to 0.8, 0.6, 0.4, 0.2, and 0.0 respectively. The visualized comparison results, as shown in the fig.~\ref{fig:ablation}, indicate that when the scale is greater than or equal to 0.4, both the Attention scale and the LoRA scale can adjust the randomness of the generated images. However, when the parameter is less than 0.4, adjusting the attention score can still increase the randomness of the generated images without degrading the image quality. In contrast, further reducing the LoRA scale leads to a drastic degradation in the quality of the generated images.
We attribute this degradation mainly to the inconsistency between training and inference. When the LoRA scale is set to 0, the model reverts to the state after the first stage training, namely the T2I inpainting model, which fails to recognize the input pattern of the reference image, thus being unable to generate reasonable images.

\section{Conclusion}

In this paper, we tackle the challenge of the background inpainting for e-commerce scenarios by introducing DreamPainter, a novel framework capable of generating visually appealing commodity images from textual prompts, with optional integration of reference images and flexible control over the similarity between target and reference images, empowering precise visual alignment with design intent. To achieve this goal, we construct a high-quality dataset, namely DreamEcom-400k, which consists of 400k diverse data pairs including commodity images, reference images, masks, and prompts. Through widely evaluations using standard metrics and subjective human evaluation, we validate that DreamPainter outperforms state-of-the-art open-source methods in both Text-to-Image (T2I) and Text-and-Reference-to-Image (TR2I) background inpainting tasks.

\clearpage
{
\small
\bibliographystyle{plain}
\bibliography{references}
}

\appendix

\section{Appendix}

\subsection{Qualitative Results on Out-of-domain Tasks.}
\label{sec:out-of-domain}

\begin{figure}[ht]
    
	\centering
	\includegraphics[width=1.0\linewidth]{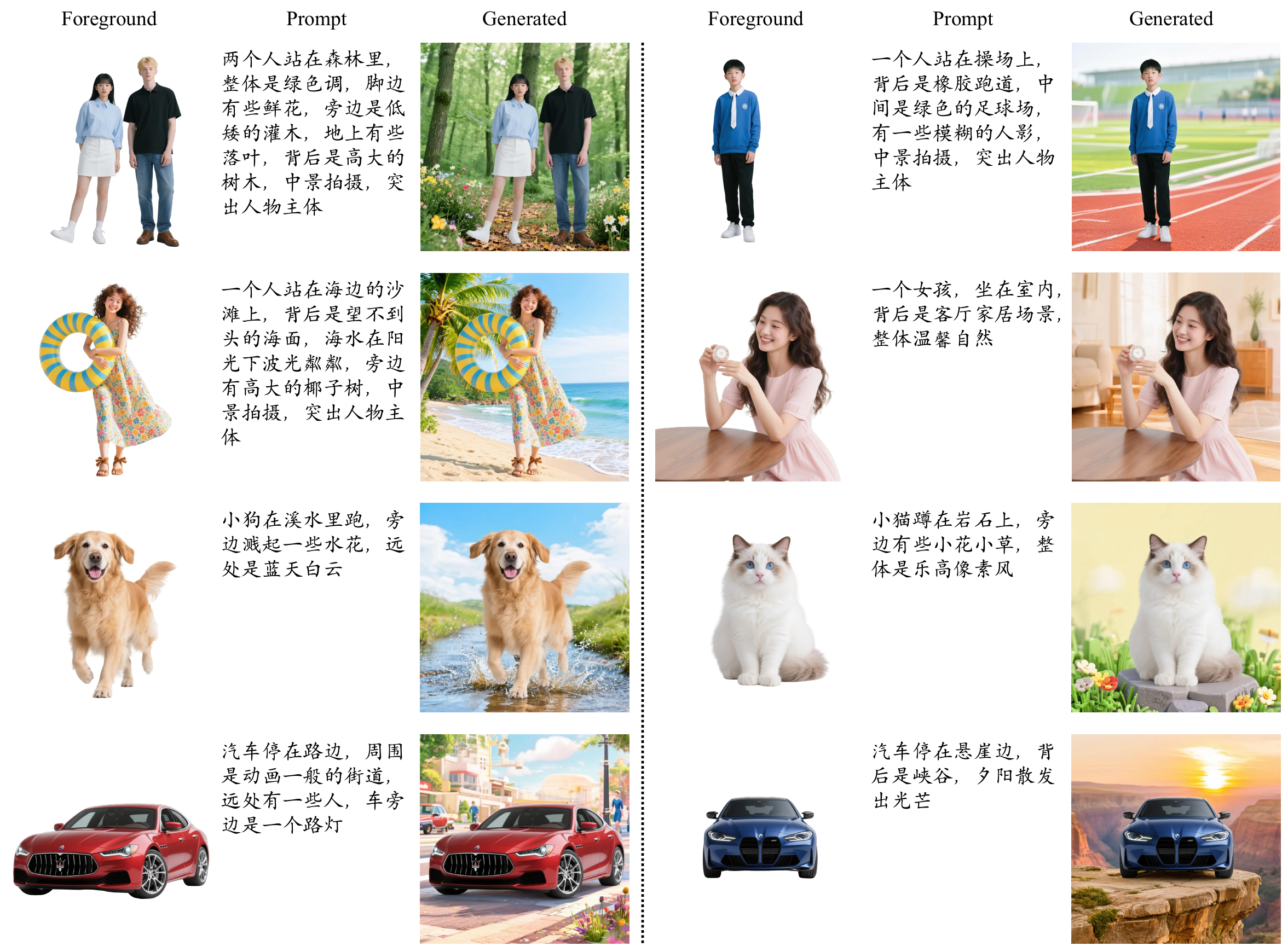}
    \caption{Qualitative results of T2I background inpainting on out-of-domain tasks.}
\label{fig:t2i_outofdomain}
\vspace{-5pt}
\end{figure}

To evaluate the generalization capability of our model in general background inpainting tasks, we applied it to generate backgrounds for scenes featuring humans, animals, and vehicles as the main subjects. As illustrated in Fig.~\ref{fig:t2i_outofdomain}, these out-of-domain tasks involve subjects distinctly different from the training data, yet the results demonstrate that our model maintains robust performance. This indicates that the model has effectively learned transferable visual representations, enabling it to generalize well to unseen subject categories and highlighting its adaptability in diverse background inpainting scenarios beyond the original training domain.

\end{document}